\pdfoutput=1
\documentclass[11pt]{article}

\usepackage{EMNLP2022}

\usepackage{times}
\usepackage{latexsym}

\usepackage{soul}
\usepackage{url}
\usepackage{hyperref}
\usepackage[utf8]{inputenc}
\usepackage{caption}
\usepackage{graphicx}
\usepackage{amsmath,amsthm,amsfonts,amssymb,mathtools}
\usepackage{booktabs}
\newcommand{\xhdr}[1]{{\noindent\bfseries #1}.}
\usepackage{multirow}
\usepackage[noend]{algpseudocode}
\usepackage{algorithm}
\usepackage{algorithmicx}
\usepackage[T1]{fontenc}
\usepackage[utf8]{inputenc}
\usepackage{microtype}
\usepackage{inconsolata}

\title{\textsc{SQUIRE}: A Sequence-to-sequence Framework for
\\ Multi-hop Knowledge Graph Reasoning}

\author{
    \textbf{Yushi Bai}$^{1,2}$, \textbf{Xin Lv}$^{1,2}$, \textbf{Juanzi Li}$^{1,2}$, \textbf{Lei Hou}$^{1,2}$\thanks{\quad Corresponding Author}, \\
    \textbf{Yincen Qu}$^{3}$, \textbf{Zelin Dai}$^{3}$, \textbf{Feiyu Xiong}$^{3}$ \\
    $^1$Department of Computer Science and Technology, BNRist; \\
    $^2$KIRC, Institute for Artificial Intelligence; \\
    Tsinghua University, Beijing 100084, China \\
    $^3$Alibaba Group, Hangzhou, China \\
    \texttt{\{bys22@mails., lijuanzi@, houlei@\}tsinghua.edu.cn}\\
    }

\begin{document}
\maketitle
\begin{abstract}

Multi-hop knowledge graph (KG) reasoning has been widely studied in recent years to provide interpretable predictions on missing links with evidential paths.
Most previous works use reinforcement learning (RL) based methods that learn to navigate the path towards the target entity.
However, these methods suffer from slow and poor convergence, and they may fail to infer a certain path when there is a missing edge along the path.
Here we present SQUIRE, the first \textbf{S}e\textbf{q}uence-to-sequence based m\textbf{u}lt\textbf{i}-hop \textbf{re}asoning framework, which utilizes an encoder-decoder Transformer structure to translate the query to a path.
Our framework brings about two benefits: 
(1) It can learn and predict in an end-to-end fashion, which gives better and faster convergence;
(2) Our transformer model does not rely on existing edges to generate the path, and has the flexibility to complete missing edges along the path, especially in sparse KGs.
Experiments on standard and sparse KGs show that our approach yields significant improvement over prior methods, while converging 4x-7x faster.

\end{abstract}

\section{Introduction}
\label{sec:intro}

Knowledge graph (KG) provides structural knowledge about entities and relations in real world in the form of triples.
Each edge in the graph, connecting two entities with a relation, represents a triple fact $(h, r, t)$.
Knowledge graph supports a variety of downstream tasks, such as question answering~\cite{hao2017end}, information retrieval~\cite{xiong2017explicit} and hierarchical reasoning~\cite{bai2021cone}.
However, practical KGs often suffer from incompleteness, thus proposing the task of KG completion, such as predicting the tail entity $t$ given $(h, r)$.
A popular approach for such a challenge is knowledge graph embedding (KGE)~\cite{bordes2013translating,dettmers2018convolutional}, which infers a missing edge in a complete black-box manner.

\begin{figure}[t]
\centering
\includegraphics[width=1\linewidth,trim=210 130 250 50,clip]{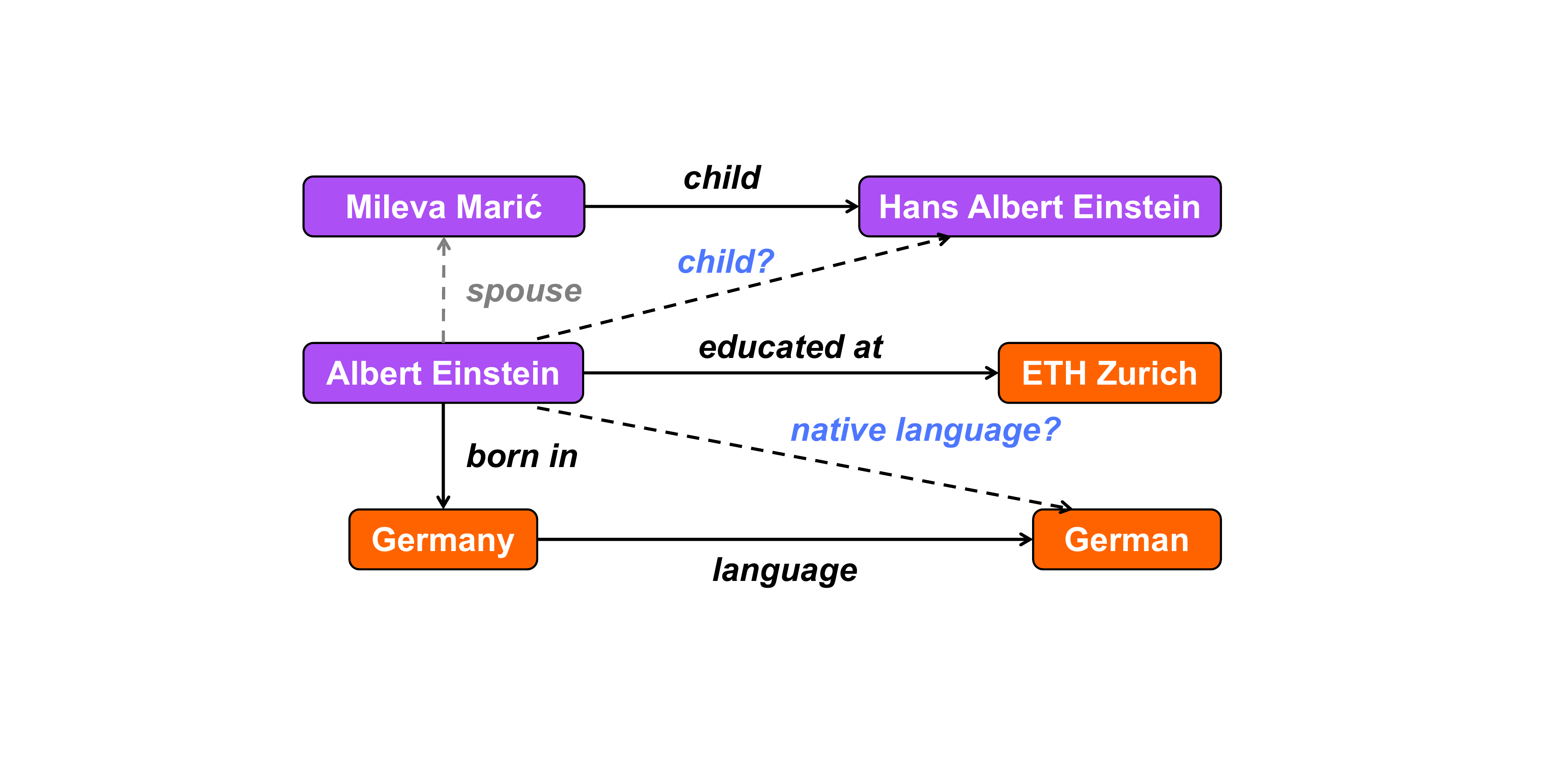} 
\caption{An example of multi-hop reasoning in an incomplete knowledge graph. The missing links (dashed arrows) can be inferred from existing links (solid arrows). However, such an evidential path may not be inferred when there is a missing edge (gray dashed arrow) along the path.}
\label{fig:example}
\end{figure}

To strengthen the interpretability of KG completion, \cite{das2018go} proposes \emph{multi-hop knowledge graph reasoning}. Given a triple query $(h, r)$, the task aims not only to predict the tail entity $t$, but to give the evidential path from $h$ to $t$ that indicates the inference process, e.g., we can infer \texttt{(Albert, native language, ?)} from the relational path ``\texttt{born in}'' and ``\texttt{language}'', as shown in Fig.~\ref{fig:example}.

\begin{figure}[t]
\centering
\includegraphics[width=1\linewidth,trim=0 0 0 0,clip]{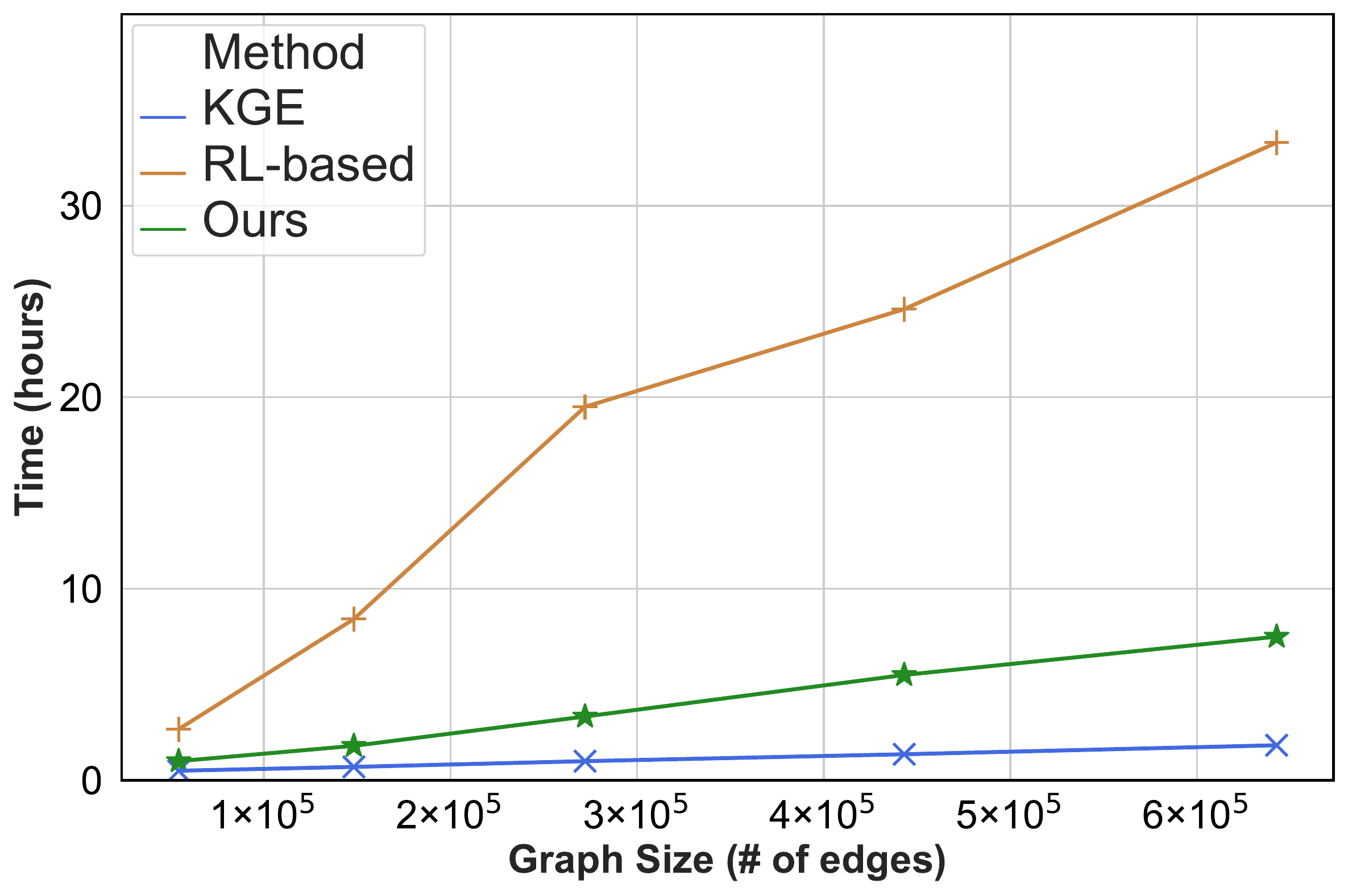} 
\caption{Training time (in hour) for KGE model, RL-based multi-hop reasoning model and our multi-hop reasoning model under varying graph sizes (measured by the number of edges).}
\label{fig:time}
\end{figure}

Most previous works use \emph{walk-based method}~\cite{das2018go,lin2018multi,lv2019adapting,lei2020learning} to tackle such problem, where an agent is trained under the reinforcement learning (RL) framework to learn to ``walk'' from the head entity to the tail entity.
One major drawback of these RL-based methods is that they suffer from slow and poor convergence, since the reward can be temporally delayed during the training process of RL~\cite{woergoetter2008reinforcement}.
In Fig.~\ref{fig:time}, we show the training time of KGE model (TransE by \citet{bordes2013translating}, blue curve) and RL-based multi-hop reasoning model (MultiHopKG by \citet{lin2018multi}, brown curve) under different graph sizes. 
Though multi-hop reasoning is a harder task than KG completion, for RL-based model, the trade-off on time is still unbearable as the size of the graph grows.
Moreover, as \citet{lv2020dynamic} points out, previous approaches suffer from the missing path problem, i.e., the model fails to infer an evidential path between pair of entities due to missing edges along the path, especially in \emph{sparse KGs}. As shown in Fig.~\ref{fig:example}, there is no evidential path between \texttt{Albert} and \texttt{Hans} due to the missing of relation ``\texttt{spouse}''.

Amazingly, we show that both drawbacks can be alleviated by a new \emph{\textbf{S}e\textbf{q}uence to sequence framework for m\textbf{u}lt\textbf{i}-hop \textbf{re}asoning} (SQUIRE).
By posing multi-hop reasoning as a sequence-to-sequence problem, we use a Transformer encoder-decoder model~\cite{vaswani2017attention} to ``translate'' a query sequence to a path sequence.

For model learning, each triple induces a supervised training sample consisting of a source sequence, i.e., query $(h, r)$, and a target path sequence sampled from all paths between $h$ and $t$.
Hence, SQUIRE framework learns and predicts in a complete end-to-end fashion, yielding faster and more stable convergence than RL-based methods (green curve in training time comparison Fig.~\ref{fig:time}). 
Furthermore, our approach naturally overcomes the missing path problem, since the Transformer does not explicitly rely on existing edges in the graph to generate the path sequence. 
That is, our proposed method has the flexibility to ``\emph{walk and complete}'': automatically infers missing edges along the path.

Meanwhile, multi-hop reasoning poses particular challenges to our framework.  \textbf{(a)} \textbf{Noisy sample}: Unlike in language modeling where we have a groundtruth target sequence, in our case, there is no such supervision on the target path.
A naive way to obtain the target path is randomly sampling from all paths between $h$ and $t$, but this might introduce noise into model learning since the random path may be spurious.
To address this, we propose \emph{rule-enhanced learning}.
We search groundtruth paths by logical rules mined from the KG, which are less noisy and more reliable compared to randomly sampled paths.
\textbf{(b)} \textbf{History gap}: During training, we provide the groundtruth sequence as the path history for our model to predict the next token.
Yet, during inference the history is generated by the model from scratch, resulting in a deviation from its training distribution.
In language modeling, this is also known as exposure bias~\cite{ranzato2015sequence}, a common challenge faced in auto-regressive models.
To narrow such a gap, we propose \emph{iterative training} that iteratively aggregates new paths to the training set based on the model's previous predictions, adapting the model to the distribution of history tokens it induces.

We evaluate the performance of SQUIRE on link prediction over six benchmark KGs.
SQUIRE achieves state-of-the-art results across all datasets.
Moreover, on two sparse KGs, our model outperforms DacKGR~\cite{lv2020dynamic}, which is specifically designed to handle the sparse setting.
SQUIRE takes \textbf{4x-7x} less time to converge on larger KG datasets while obtaining better performance.
To the best of our knowledge, SQUIRE is the first sequence-to-sequence framework for multi-hop reasoning, and may provide a new paradigm for future studies.

\section{Related Work}
\label{sec:related}

\begin{figure*}[t]
\centering
\includegraphics[width=0.95\linewidth,trim=40 40 40 25,clip]{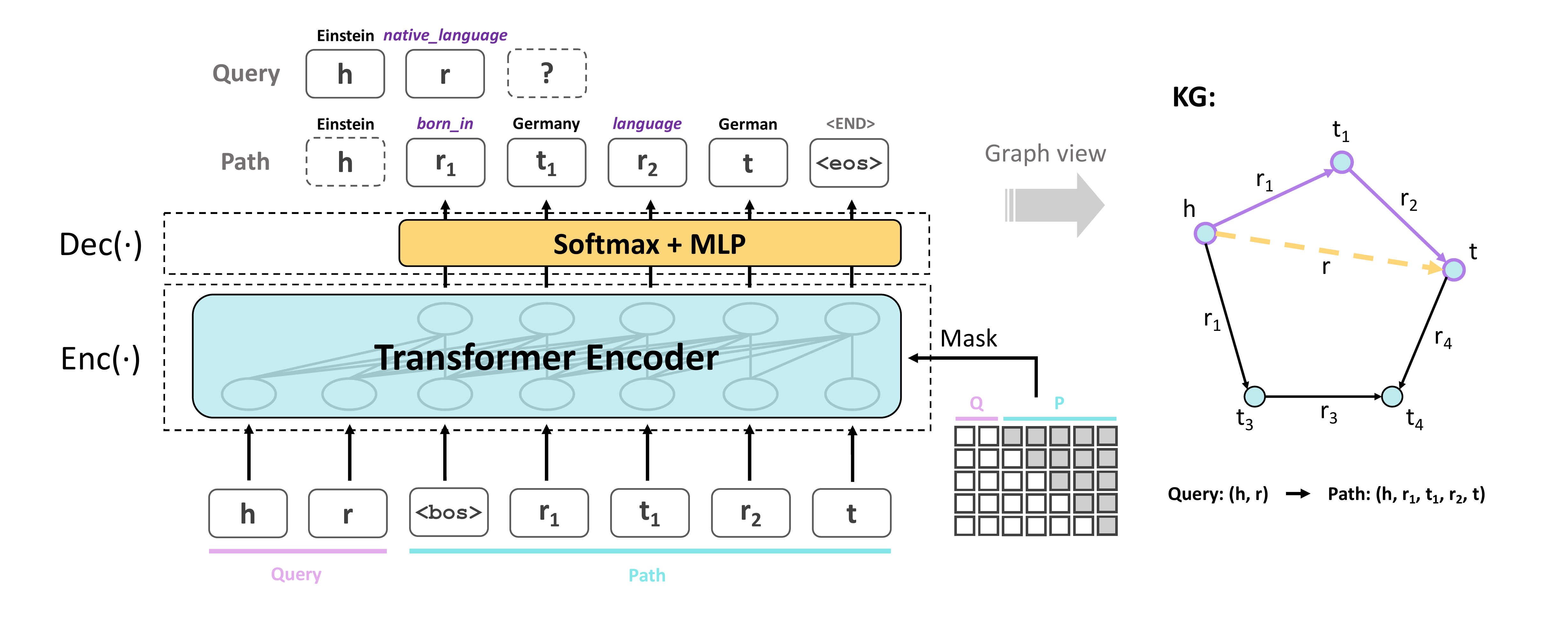}
\caption{SQUIRE model overview: We use Transformer encoder to compute a contextualized representation from query tokens and history path tokens (the mask makes sure only previous tokens in the path can be attended to), and decode the next token through MLP and Softmax layer. The generated sequence corresponds to a path in KG.}
\label{fig:model}
\end{figure*}

\subsection{Knowledge Graph Embedding}
Knowledge graph embedding (KGE) methods map entities to vectors in low-dimensional embedding space, and model relations as transformations between entity embeddings.
Prominent examples include TransE~\cite{bordes2013translating}, ConvE~\cite{dettmers2018convolutional}, RotatE~\cite{sun2019rotate} and TuckER~\cite{balavzevic2019tucker}.
Each of these models is equipped with a scoring function that maps any triple to a scalar score, which measures the likelihood of the triple.
The embeddings of entities and relations are learnt by optimizing the scoring function such that the likelihood score is high for true triples while low for false triples.

\subsection{Multi-hop Reasoning}
Multi-hop reasoning aims not only on finding the target entity to the query $(h, r, ?)$, but also the reasoning path from $h$ to $t$ to support the prediction.

DeepPath~\cite{xiong2017deeppath} is the first approach to adopt an RL framework for multi-hop reasoning.
Following this work, MINERVA~\cite{das2018go} introduces the REINFORCE algorithm to this task.
Considering the incomplete KG environment that might give low-quality rewards, MultiHopKG~\cite{lin2018multi} proposes reward shaping from a pretrained KGE model.
DacKGR~\cite{lv2020dynamic} further applies dynamic anticipation and completion in sparse KGs.

In addition to the above RL-based approaches, several symbolic rule-based models are proposed to improve interpretability for KG completion, including NTP~\cite{rocktaschel2017end}, NeuralLP~\cite{yang2017differentiable} and AnyBURL~\cite{chris2019anyburl}.
Most of these methods offer evidence from automatically learnt logical rules, which are, to some degree, weaker in interpretability than RL-based methods that give the evidential path.

\subsection{Reinforcement Learning via Transformer}
Interestingly, recent works~\cite{chen2021decision,janner2021offline} have shown the feasibility to treat RL as a sequence modeling problem.
They use Transformer to model the trajectory including states, actions and rewards, and their architectures achieve promising results on offline RL tasks.
On our multi-hop reasoning task, these findings suggest the potential of substituting the previous RL pipeline with Transformer.

\section{Methodology}
\label{sec:model}

\subsection{Preliminaries}
\xhdr{Knowledge graph}
We denote the set of entities and relations in knowledge graph as $\mathcal{E}$ and $\mathcal{R}$.
Each directed edge in the graph can be formalized as a triple $(h, r, t)$, where $h, t\in\mathcal{E}$ and $r\in\mathcal{R}$. 
Let $\mathcal{T}$ denote the set of all such triple facts.

\xhdr{Multi-hop reasoning}
Given triple query $(h, r, ?)$, multi-hop reasoning aims to find an evidential path $h, r_1, t_1, \dots, r_n, t$ towards the target entity $t$, where $t_i$ is the intermediate entity along the path connected by relational edges $r_i$ and $r_{i+1}$. $n$ suggests that it is an $n$-hop path, that is, the length of the path is $n$.

\subsection{SQUIRE Framework}
Our end-to-end SQUIRE approach frames multi-hop reasoning as a sequence-to-sequence task.
The query is treated as the source sequence and the path as the target sequence.
As shown in Fig.~\ref{fig:model}, we utilize a Transformer encoder to map query and previous path sequence to a contextualized representation, and further use such representation to autoregressively decode the output path, token by token.

The intuition behind our framework is simple: multi-hop reasoning task setting resembles the question-answering task in NLP: different entities and relations can be interpreted as different words whose contextualized embeddings can be learnt from edges in the graph.
Furthermore, relational rules like $r\rightarrow r_1\land r_2$ (e.g., ``\texttt{native language} $\rightarrow$ \texttt{born in} $\land$ \texttt{language}'' in Fig.~\ref{fig:example}) can be learnt from training samples where $r$ in the query is decomposed into $r_1, r_2$ in the path.

To illustrate the training and inference process of SQUIRE, let $q, \tau$ denote the query and the path: 
\begin{align}
    q := (h, r),\ \tau := (r_1, t_1, \dots, r_n, t, \text{\textless eos\textgreater})
\end{align}
and $\tau_k$ denotes the $k$-th token in the output path. We decompose the probability
\begin{equation}
    p\left(\tau\, |\, q\right) = \prod_{k=1}^{|\tau|}p\left(\tau_k\,|\,q, \tau_{<k}\right)
\end{equation}
where $|\tau|$ is the number of tokens in the path.

The model learns a token embedding matrix $\mathbf{E}\in \mathbb{R}^{V\times d}$ under embedding dimension of $d$. The vocabulary size $V=|\mathcal{E}|+|\mathcal{R}|+|\mathcal{S}|$, since the token vocabulary includes the set of entities $\mathcal{E}$, the set of relations $\mathcal{R}$ and the set of special tokens $\mathcal{S}$.\footnote{Special tokens include start (\textless bos\textgreater ), end (\textless eos\textgreater ) and mask (\textless mask\textgreater ).}
We use $\text{Enc}(\cdot)$ to denote the encoder in our model (marked in Fig.~\ref{fig:model}). 
When decoding $\tau_k$, the mask, shown in Fig.~\ref{fig:model}, only allows the encoder to attend to the query tokens $h, r$ and previous tokens $\tau_{<k}$ in the output path, preventing the revelation
of future information.
Then our model computes the probability distribution on the $k$-th token as
\begin{equation}
\begin{aligned}
    p\left(\cdot | q, \tau_{<k}\right) 
    = \text{Softmax}(\text{MLP}(\text{Enc}(h, r, \tau_{<k}))\cdot \mathbf{E})
\end{aligned}
\end{equation}
where the $\text{MLP}(\cdot)$ is a multi-layer perceptron that learns a mapping $\mathbb{R}^d\rightarrow\mathbb{R}^d$.

During training, for each triple fact $(h, r, t)\in\mathcal{T}$, we sample a path from all paths with maximum length $N=3$~\footnote{We choose such $N$ since almost all pairs of entities are within 3 hops in the KG datasets, and to keep fair comparison with baselines that all set maximum path length to 3.} between $h$ and $t$ and treat it as the target path $\tau$.
If such a path cannot be found, we simply take $\tau:=(r, t, \text{\textless eos\textgreater })$ to force the model to memorize the edge.
After obtaining the set $\mathcal{U}$ of all query-path pairs, 
we optimize the parameters to maximize $\sum_{(q, \tau)\in\mathcal{U}}p\left(\tau\, |\, q\right)$. 
We use cross-entropy loss and further add label-smoothing to avoid overfitting, resulting in the following loss function for each sample $(q, \tau)\in \mathcal{U}$:
\begin{equation}
    \mathcal{L} = -\frac{1}{|\tau|}\sum_{k=1}^{|\tau|}\sum_{i=1}^{V}\alpha_i\, \log p\left(i\, |\,q, \tau_{<k}\right)
\end{equation}
where $\alpha_i=\epsilon$ for the target token $i=\tau_k$ and $\alpha_i=\frac{1-\epsilon}{V-1}$ for other tokens.
$\epsilon$ is the label-smoothing hyperparameter ranging from 0 to 1.
Furthermore, to avoid overdependence of future predictions on path history, we mask out (by substituting with \textless mask\textgreater \, token) every entity token in $\tau$ with probability $p$, and exclude the loss on masked tokens.

For multi-hop KG reasoning task, given query $(h, r, ?)$, we use beam search to generate reasoning path $\tau^*$ of maximum length $N$ from $q=(h, r)$:
\begin{equation}
    \tau^*=\arg\max_{\tau}\frac{1}{|\tau|}\sum_{k=1}^{|\tau|}\log p\left(\tau_k\, |\, q, \tau_{<k}\right)
\label{eq:beam}
\end{equation}

Our sequence-to-sequence framework brings two challenges.
\textbf{(a)} \emph{Noisy sample}: Notice that during training, we sample path from $h$ to $t$ as the target path, for there is no golden standard for a ``groundtruth'' reasoning path. This might result in low-quality paths and introduce noise into model learning.
\textbf{(b)} \emph{History gap}: There is a gap between the history path tokens during training and inference, where the former is drawn from groundtruth data distribution while the latter is induced by the model.
Specifically, for a sample $(q, \tau)$, the model is trained to maximize $p\left(\tau_k\, |\, q, \tau_{<k}\right)$ on the $k$-th token. However, during inference, the probability distribution on the $k$-th token, $p\left(\cdot\, |\, q, \tau'_{<k}\right)$, is modeled based on $\tau'_{<k}$ that is generated by the model.
The distribution of $\tau'_{<k}$ may deviate from groundtruth distribution of $\tau_{<k}$ during training, and the gap may even be larger with a longer history sequence, eventually leading to a false target entity during inference.
To address these challenges, we propose \emph{rule-enhanced learning} and \emph{iterative training}, as described in detail in the following sections.

\subsection{Rule-enhanced Learning}
To address the noisy sample challenge, we propose rule-enhanced learning that uses mined rules to guide path searching and obtain high quality path-query pairs.
Inspired by the recent advances in rule-based multi-hop reasoning methods~\cite{yang2017differentiable,sadeghian2019drum}, we utilize AnyBURL~\cite{chris2019anyburl}, which is the Sota method for rule mining on KG, to efficiently mine logical rules. 
Each rule decomposes a single relation into the composition of multiple relations (including inverse relations).\footnote{These rules are a subset of rules in \cite{chris2019anyburl}, referred to as \textbf{C} rules in their paper.}
A confidence score is given along with each rule, and we choose the rules with confidence scores larger than some threshold and treat them as ``golden rules''.
Then we use these rules to find the evidential path between $h$ and $t$.
For example, if one rule for relation $r$ is $r(X, Y)\rightarrow r_1(X, A_1)\land r_2(A_1, A_2)\land \dots\land r_n(A_{n-1}, Y)$, we traverse the path from $h$ to $t$ along the relational edges $r_1, r_2, \dots, r_n$.
If none of the rules lead to a valid path in the graph, we obtain the path by random sampling.

However, utilizing rule-based method to generate valid query-path pairs introduces the noisy rule problem.
This is because some relational rules do not hold for all entities, and may lead to unreasonable paths from $h$ to $t$ for certain entities.
We show that an iterative training strategy, elaborated in the next section, can alleviate the noisy rule problem.

\begin{algorithm}[t]
\small
\begin{algorithmic}[1]
\State $\mathcal{T}\leftarrow$ set of all triples in the graph
\State Initialize query-path training set $\mathcal{U}$ by random sampling or rule-enhanced searching
\State Initialize model $M$
\State Train $M$ for $n$ epochs
\For{$k=2$ to $N$}
\For{each triple $(h, r, t)$ in $\mathcal{T}$}
\State $q\leftarrow (h, r)$
\State $\tau_1 \leftarrow$ first $(k-1)$ hops of $M(q)$
\State $\tau_2 \leftarrow$ subsequent path of $\tau_1$ towards $t$, at most
\Statex \qquad \qquad \qquad $(N-k+1)$ hops
\If{$\tau_2\ne null$}
\State $\tau \leftarrow \tau_1 + \tau_2$
\Else \Comment{No valid subsequent path}
\State $\tau\leftarrow$ search for an entire path from $h$ to $t$
\EndIf
\State Add $(q, \tau)$ to $\mathcal{U}$
\EndFor
\State Train $M$ for $n/k$ epochs
\EndFor
\State \textbf{Return} trained model $M$

\end{algorithmic}
\caption{Iterative Training}
\label{alg:iter}
\end{algorithm}

\subsection{Iterative Training}
Our history gap challenge is also a commonly faced problem in autoregressive models (known as exposure bias in language modeling), where the core idea behind the solutions~\cite{venkatraman2015improving,zhang2019bridging} is to train the model to predict under the same condition during inference.
With similar intuition in mind, we propose a novel training strategy that iteratively aggregates new training data based on the model's prediction to help the model adapt to its induced distribution of history tokens.
Our data aggregation idea is enlightened by the DAgger algorithm~\cite{ross2011reduction}, which is designed to boost RL performance in sequential prediction problems.

Our algorithm proceeds as follows.
At the first iteration, we generate query-path training set $\mathcal{U}$ as illustrated before and train the model for several epochs.
At $k$-th iteration ($k>1$), for each triple in the graph, we partially leverage the current model to find the path and add a new training sample to $\mathcal{U}$.
Specifically, our algorithm uses the current model to predict the first $(k-1)$-hops, and search the path following these tokens with maximum length $N$.
If the subsequent path cannot be found, it may be due to the model's failure on predicting the first $(k-1)$-hops, then we search the entire path again to strengthen the model's learning on this sample.
After data aggregation at $k$-th iteration, the size of the training set becomes $k$ times the initial size, and we continue to train the model on the new training set for the same number of steps before the next iteration.
The total number of iterations would be $N$, which is the maximum hops of the path.
A detailed algorithm is shown in Alg.~\ref{alg:iter}.

Note that during $k$-th iteration, $\tau_2$ is randomly sampled after $\tau_1$, one may worry that this brings noise into training data. However, since the model has learnt ``soft'' rules in the past iterations, making $\tau_1$ much more reasonable than random sampling, then the search space left for $\tau_2$ is more concentrated around the groundtruth path.

Additionally, the iterative training strategy can mitigate the noisy rule problem introduced by rule-enhanced learning. The paths obtained by such noisy rules may be replaced during the data aggregation step, meanwhile, paths that entail our model's previous predictions may serve as more solid training samples.

\section{Experiments}
\label{sec:experiments}

\begin{table}[t]
\centering
\resizebox{0.49\textwidth}{!}{
\begin{tabular}{lrrrrr}
\toprule[2pt]
\multirow{2}{*}{\textbf{Dataset}} & \multirow{2}{*}{\textbf{\#Ent}} & \multirow{2}{*}{\textbf{\#Rel}} & \multirow{2}{*}{\textbf{\#Fact}} & \multicolumn{2}{c}{\textbf{\#Degree}} \\ \cmidrule(l){5-6} 
                         &                        &                        &                         & \textbf{mean}         & \textbf{median}        \\ \midrule
FB15K237                & 14,505                 & 237                    & 272,115                 & 18.71        & 13            \\
NELL995                  & 62,706                 & 198                    & 117,937                 & 1.88         & 1             \\
FB15K237-20\%           & 13,166                 & 237                    & 54,423                  & 4.13         & 3             \\
NELL23K                  & 22,925                 & 200                    & 25,445                  & 1.11         & 1             \\
KACC-M & 99,615 & 209 & 642,650 & 6.45 & 4 \\
FB100K & 100,030 & 471 & 1,013,470 & 10.13 & 7
\\ \bottomrule[2pt]
\end{tabular}
}
\caption{Dataset statistics.}
\label{tb:dataset}
\end{table}

\begin{table*}[t]
\centering  
\resizebox{\textwidth}{!}{
\begin{tabular}{lcccccccccccccccc}
\toprule[2pt]
  & \multicolumn{4}{c}{FB15K237} & \multicolumn{4}{c}{NELL995} & \multicolumn{4}{c}{FB15K237-20\%} & \multicolumn{4}{c}{NELL23K} \\
\cmidrule(l){2-5} \cmidrule(l){6-9} \cmidrule(l){10-13} \cmidrule(l){14-17}
& MRR & H@1 & H@3 & H@10 & MRR & H@1 & H@3 & H@10 & MRR & H@1 & H@3 & H@10 & MRR & H@1 & H@3 & H@10 \\
\midrule
\emph{Embedding-based methods} \\
TransE~\cite{bordes2013translating} & .425 & .320  & .475 & .635 & .371 & .209 & .473 & .654 & .263 & .178 & .288 & .434 & .179 & .076 & .208 & .379 \\  
ConvE~\cite{dettmers2018convolutional} & .438 & .342 & .483 & .627 & .542 & \underline{.449} & .594 & .709 & .264 & .187 & .284 & .422 & \underline{.279} & \underline{.193} & \underline{.301} & \underline{.467} \\ 
RotatE~\cite{sun2019rotate} & .426 & .321 & .474 & .635 & .513 & .411 & .570 & .708 & .265 & .185 & .286 & .430 & .217 & .141 & .232 & .368 \\
TuckER~\cite{balavzevic2019tucker} & \underline{.451} & \underline{.357} & \underline{.495} & .635 & .511 & .422 & .556 & .682 & .246 & .178 & .261 & .384 & .207 & .143 & .224 & .338 \\
ConE~\cite{bai2021cone} & .446 & .345 & .490 & \underline{.645} & \underline{.543} & .448 & \underline{.602} & \underline{.715} & \underline{.274} & \underline{.193} & \underline{.298} & \underline{.437} & .234 & .158 & .249 & .400 \\
\midrule
\emph{Rule-based methods} \\
AnyBURL~\cite{chris2019anyburl} & - & .300 & .405 & .544 & - & .389 & .521 & .628 & - & .159 & .240 & .359 & - & .140 & .203 & .292 \\
\midrule
\emph{RL-based methods} \\
MINERVA~\cite{das2018go} & .275 & .199 & .306 & .433 & .391 & .293 & .449 & .575 & .123 & .070 & .133 & .236 & .151 & .101 & .159 & .247 \\
MultiHopKG~\cite{lin2018multi} & \underline{.407} & \underline{.327} & \underline{.443} & \underline{.564} & \underline{.467} & \underline{.388} & \underline{.512} & \underline{.609} & .231 & .167 & .250 & .361 & .178 & .124 & .188 & .297 \\
RuleGuider~\cite{lei2020learning} & .387 & .297 & .428 & .563 & .417 & .344 & .476 & .582 & .094 & .042 & .094 & .21 & .112 & .030 & .140 & .273 \\
DacKGR~\cite{lv2020dynamic} & .347 & .274 & .382 & .493 & .421 & .347 & .464 & .554 & \underline{.246} & \underline{.180} & \underline{.270} & \underline{.386} & \underline{.197} & \underline{.133} & \underline{.211} & \underline{.337} \\
\midrule
\emph{Sequence-based methods} \\
SQUIRE & .421 & .329 & .465 & .606 & .498 & .409 & .550 & .668 & .249 & .180 & .272 & .401 & .233 & .157 & .256 & .389 \\
SQUIRE + Self-consistency & {\bf .433} & {\bf .341} & {\bf .476} & {\bf .617} & {\bf .519} & {\bf .434} & {\bf .570} & {\bf .682} & {\bf .253} & {\bf .180} & {\bf .276} & {\bf .406} & {\bf .244} & {\bf .165} & {\bf .269} & {\bf .412} \\
\bottomrule[2pt]
\end{tabular}
}
\caption{Link prediction results on four benchmark datasets. The best score of multi-hop reasoning models is in \textbf{bold}. The best score among embedding-based models and the best score among RL-based models are \underline{underlined}.}
\label{tb:link}
\end{table*}

\subsection{Experimental Setup}
\xhdr{Datasets}
We experiment on six KG benchmarks that capture common knowledge. Two standard datasets include FB15K237~\cite{toutanova2015observed} which is extracted from Freebase, and NELL995\footnote{We apply a new training/valid/test split on the whole NELL995 graph (without inverse relations), since there is an inconsistency in evaluation in previous studies.}~\cite{xiong2017deeppath} that is constructed from NELL. Two sparse datasets include FB15K237-20\%~\cite{lv2020dynamic} which is constructed by randomly retaining 20\% triples in the training set of FB15K237, and NELL23K~\cite{lv2020dynamic}, constructed by randomly sampling a small proportion of edges in a subgraph of NELL. 
For efficiency studies, we also consider two larger KGs, KACC-M~\cite{zhou2021kacc} that is constructed based on Wikidata, and FB100K that we extract from Freebase.
Detailed statistics of the four datasets are listed in Table~\ref{tb:dataset}.\footnote{NELL995 is, in fact, sparse (according to the statistic in Table~\ref{tb:dataset}), but we consider it a standard KG benchmark as previous multi-hop reasoning studies do.}

\xhdr{Baselines}
For embedding-based models, we compare with TransE~\cite{bordes2013translating}, ConvE~\cite{dettmers2018convolutional}, RotatE~\cite{sun2019rotate}, TuckER~\cite{balavzevic2019tucker} and ConE~\cite{bai2021cone}. For multi-hop reasoning models, we take MINERVA~\cite{das2018go}, MultiHopKG~\cite{lin2018multi}, RuleGuider~\cite{lei2020learning} and DacKGR~\cite{lv2020dynamic} as baseline models.~\footnote{More recent baselines RLH~\cite{wan2021reasoning}, RARL~\cite{hou2021rule} have not released their code and we fail to reproduce their results.}
For rule-based reasoning models, we compare with AnyBURL\footnote{During inference, we only include C rules mined by AnyBURL, as our model only utilize these rules during rule-enhanced learning.}~\cite{chris2019anyburl}. 

\xhdr{Evaluation Protocol}
We follow the evaluation protocol in most multi-hop reasoning works~\cite{das2018go,lin2018multi}. For every triple $(h, r, t)$ in the test set, we convert it to a triple query $(h, r, ?)$ and obtain a ranking list of the tail entity from the model.
We compute the metrics, including Mean Reciprocal Rank (MRR) and Hits at N (H@N) under filtered setting~\cite{bordes2013translating}.

\xhdr{Implementation Details}
We use Adam~\cite{kingma2014adam} as the optimizer to train our model. We search hyperparameters including batch size, embedding dimension, learning rate, label smoothing factor, mask probability and warmup steps (see training details and best hyperparameters in Appendix~\ref{sec:detail}).\footnote{The code of our paper is available at \url{https://github.com/bys0318/SQUIRE}.}
During path sampling and generation, we also involve inverse edges (inverse relations are also added to the set $\mathcal{R}$ of all relations), thus each training triple induces two edges of opposite directions. 
During evaluation, we perform beam search to obtain a list of multi-hop paths, along with log-likelihood as their scores.
To prevent the model from giving higher scores to shorter paths, we further divide the log-likelihood score by the length of the path, and obtain a final score for each path (as shown in Eq.~\ref{eq:beam}).
Finally, we sort the target entities according to the maximum score among all paths that lead to them.
In addition, we consider adding \emph{self-consistency}~\cite{wang2022self} into decoding the target entity. The score for each entity is the summation of the probability of all generated paths that lead to it, and we sort the target entities according to their total probability.

\begin{table}[t]
\centering  
\resizebox{0.49\textwidth}{!}{
\begin{tabular}{lcccccccc}
\toprule[2pt]
 & \multicolumn{4}{c}{FB15K237} & \multicolumn{4}{c}{NELL995} \\
\cmidrule(l){2-5} \cmidrule(l){6-9}
& MRR & H@1 & H@3 & H@10 & MRR & H@1 & H@3 & H@10 \\
\midrule
SQUIRE & {\bf .421} & {\bf .329} & {\bf .465} & {\bf .606} & {\bf .498} & {\bf .409} & {\bf .550} & {\bf .668} \\  
-iter & .411 & .321 & .452 & .594 & .488 & .402 & .536 & .654 \\ 
-iter -rule & .386 & .294 & .430 & .572 & .482 & .394 & .532 & .653 \\
\bottomrule[2pt]
\end{tabular}
}
\caption{Ablation study on our proposed iterative training strategy and rule-enhanced learning. ``-iter'' refers to the model without iterative training, ``-iter -rule'' refers to one that further removes rule-enhanced learning.}
\label{tb:ablation}
\end{table}

\subsection{Results}
Table~\ref{tb:link} reports the link prediction results on the first four datasets.
We observe that KGE models attain better results across four datasets, while the gaps on MRR metric between the best KGE results and the best multi-hop results are all smaller than 5\%.
Hence, it is worth sacrificing a little performance for interpretability on triple query answering.

Among previous RL-based methods, MultiHopKG performs superior to others on standard datasets (left two). DacKGR provides the most precise inference on sparse datasets (right two), since it is specially designed to handle sparse setting, yet it does not perform well in the standard KGs.
We observe that our model outperforms all previous multi-hop reasoning methods across four datasets by a large margin on most metrics, regardless of the sparsity of the graph and SQUIRE does not require additional design as DacKGR does.
Also, adding self-consistency during decoding can further boost SQUIRE's performance for 1\%$\sim$2\% on all metrics. Note that we do not add self-consistency in the following studies.

The performance gain of SQUIRE on sparse KG is due to the flexibility of our framework, allowing the model to dynamically complete the graph while generating the path (a more detailed analysis is in Sec.~\ref{sec:sparse}).
We further study how the Transformer model in SQUIRE infers the evidential path by attention visualization. 
The visualization result suggests that the Transformer model has memorized the graph during training, and in generation phase it predicts the next token based on the current position and adjacent nodes or edges (see results and detailed analysis in Appendix~\ref{sec:att}). Thus we don't need to provide any external information about the graph to our model during inference.

\subsection{Ablation Studies}
We present the ablation studies on our proposed iterative training strategy and rule-enhanced learning in Table~\ref{tb:ablation}.
We can see that both techniques are beneficial to the overall performance of SQUIRE.
As the numbers suggest, the two strategies play a more important role on FB15K237 than on NELL995. 
The reason why SQUIRE benefits more on FB15K237 lies in the density of the two graphs: FB15K237 is denser and thus the two strategies can help distinguish the real evidential path among a larger set of random paths between $h$ and $t$, with the help of mined rules or trained model.
Also observe that SQUIRE with rule-enhanced learning significantly outperforms AnyBURL, indicating that our model is learning from the paths, rather than fitting the mined rules by AnyBURL.

To obtain a deeper understanding of how the two strategies help during the training process of SQUIRE, we compare the convergence rate of the original model with the two ablated models.
The convergence analysis indicates (see detailed analysis in Appendix~\ref{sec:convergence}): (a) New aggregated data during training helps the overall performance of the model; 
(b) Rule-guided searching improves the quality of training paths, resulting in a more stable convergence.

\subsection{Efficiency Studies on Larger KG}

\begin{table}[t]
\centering  
\resizebox{0.49\textwidth}{!}{
\begin{tabular}{lcccc}
\toprule[2pt]
Dataset & FB15K237 & NELL995 & KACC-M & FB100K \\
\midrule
MINERVA & 18.4 & 11.6 & 32.4 & 55.3 \\
MultiHopKG & 19.5 & 12.0 & 33.3 & 57.8 \\
SQUIRE & 3.2 (\textbf{6x}) & 3.5 (\textbf{4x}) & 7.5 (\textbf{4x}) & 8.5 (\textbf{7x}) \\
\bottomrule[2pt]
\end{tabular}
}
\caption{Training time (in hour) of RL-based MINERVA, MultiHopKG and our model on four datasets including two standard KGs and two larger KGs. All models are trained on one RTX 3090 GPU.}
\label{tb:time}
\end{table}

\begin{table}[t]
\centering  
\resizebox{0.49\textwidth}{!}{
\begin{tabular}{lcccccc}
\toprule[2pt]
 & \multicolumn{3}{c}{KACC-M} & \multicolumn{3}{c}{FB100K} \\
\cmidrule(l){2-4} \cmidrule(l){5-7}
& MRR & H@1 & H@10 & MRR & H@1 & H@10 \\
\midrule
MultiHopKG & .576 & .492 & {\bf .713} & .652 & {\bf .601} & .744 \\  
SQUIRE & {\bf .578} & {\bf .515} & .702 & {\bf .655} & .595 & {\bf .766} \\ 
\bottomrule[2pt]
\end{tabular}
}
\caption{Link prediction results on two larger KGs.}
\label{tb:large}
\end{table}

\begin{table*}[t]
\centering
\resizebox{0.75\textwidth}{!}{
\begin{tabular}{c|l}
\toprule[2pt]
\multirow{4}{*}{\begin{tabular}[c]{@{}c@{}}w/o\\ missing\end{tabular}} 
& Query: \textit{(Lycoming County, currency, ?)} \\
& Path: \textit{Lycoming County} $\xleftarrow{\textit{contains}}$ \textit{State of Pennsylvania} $\xrightarrow{\textit{currency}}$ \textit{\underline{US Dollar}} \\ \cmidrule(l){2-2} 
& Query: \textit{(Gale Sayers, won trophy, ?)} \\
& Path: \textit{Gale Sayers} $\xrightarrow{\textit{play for}}$ \textit{Bears} $\xleftarrow{\textit{coach team}}$ \textit{Urlacher} $\xrightarrow{\textit{won trophy}}$ \textit{\underline{Super Bowl}} \\ \midrule
\multirow{4}{*}{\begin{tabular}[c]{@{}c@{}}w/\\ missing\end{tabular}}  
& Query: \textit{(Vladimir Guerrero, athlete home stadium, ?)} \\
& Path: \textit{Vladimir Guerrero} $\xrightarrow{\textit{\textbf{play for}}}$ \textit{Anaheim Angels} $\xrightarrow{\textit{team home stadium}}$ \textit{\underline{Edison Field}} \\ \cmidrule(l){2-2} 
& Query: \textit{(Haifeng Xu, work for, ?)} \\
& Path: \textit{Haifeng Xu} $\xrightarrow{\textit{\textbf{lead organization}}}$ \textit{\underline{Chinese National Shooting Team}} \\ 
\bottomrule[2pt]
\end{tabular}
}
\caption{Case study of SQUIRE on link prediction. We show each triple query along with our model's predicted evidential path, the correct tail entities are \underline{underlined}. ``w/o missing'' indicates the predicted paths contain only edges in the graph, while ``w/ missing'' suggests the predicted paths contain valid edges missing from the graph due to incompleteness (missing edges in \textbf{bold}).}
\label{tb:case}
\end{table*}

\begin{table}[t]
\centering  
\resizebox{0.48\textwidth}{!}{
\begin{tabular}{lccc}
\toprule[2pt]
Model & MINERVA & MultiHopKG & SQUIRE \\
\midrule
Interpretability score & 12.5 & 15.6 & \textbf{20.8} \\
Reasonable rate (\%) & 2.1 & 2.1 & \textbf{7.3} \\
\bottomrule[2pt]
\end{tabular}
}
\caption{Interpretability evaluation results based on manual annotation. The interpretability score is the average score multiplied by 100, and the reasonable rate measures the ratio of reasonable generated paths (paths that are scored 1).}
\label{tb:interpretability}
\end{table}

To provide empirical evidence for the efficiency of our framework, we report the training time of SQUIRE and RL-based baselines on two standard KGs and two larger KGs (Table~\ref{tb:time}).
The link prediction results on two larger KGs are shown in Table~\ref{tb:large}, where we select MultiHopKG as the baseline since it performs consistently well on standard KGs.

We can see that our sequence-to-sequence framework for multi-hop reasoning brings 4x-7x speed-up across the four datasets.
Typically, MultiHopKG model takes more than a day to train on the two larger KGs, while our method converges within several hours while obtaining comparable performance.
Note that SQUIRE's performance on two larger datasets (KACC-M and FB100K) is conducted without iterative training, as we find out that iterative training on larger datasets may be time costing. As shown in Table~\ref{tb:large}, even without iterative training, SQUIRE achieves comparable performance. SQUIRE's performance can be further boosted on larger KGs with iterative training but with a trade-off for efficiency.

\subsection{Interpretability Evaluation}

To show SQUIRE can generate interpretable paths given triple queries, we provide case studies on link prediction in Table~\ref{tb:case}.
We give four examples of triple query along with the Hits@1 inferred evidential path, including paths containing only existing edges (w/o missing) and paths containing missing edges (w/ missing).
From the listed cases, we see that SQUIRE can provide reasonable paths, and it can dynamically complete the path during generation, thus yielding superior performance on sparse datasets.

Moreover, we manually annotate the interpretability score for paths generated by our model and baseline models (MINERVA and MultiHopKG).
For each triple query, we select the top generated reasoning path that leads to the correct tail entity and score it based on whether it is convincing to a human.
Following \citet{lv2021multi}, we give 1, 0.5 and 0 scores for paths that are reasonable, partially reasonable and unreasonable respectively (see detailed annotation rules and examples in Appendix~\ref{sec:case}).
We report the evaluation result on FB15K237 in Table~\ref{tb:interpretability}.
We observe that SQUIRE achieves higher scores on both metrics~\footnote{The low reasonable rate ($<10\%$ for all models) is partially due to the non-existence of such reasonable path in the incomplete KG dataset, as we observed during annotation.}. 
This suggests that our model can generate more reasonable paths and thus lead towards explainable multi-hop reasoning.

\subsection{No-constraint Inference in Sparse Setting}
\label{sec:sparse}
Earlier in the paper, we suggest our performance gain on sparse graphs comes from no-constraint generation, where there is no constraint on the path sequence generated by the model.
It gives the model the flexibility to dynamically complete the path during generation.
Here we show empirical evidence to support such a statement.

\begin{figure}[t]
\centering
\includegraphics[width=0.9\linewidth,trim=0 0 0 0,clip]{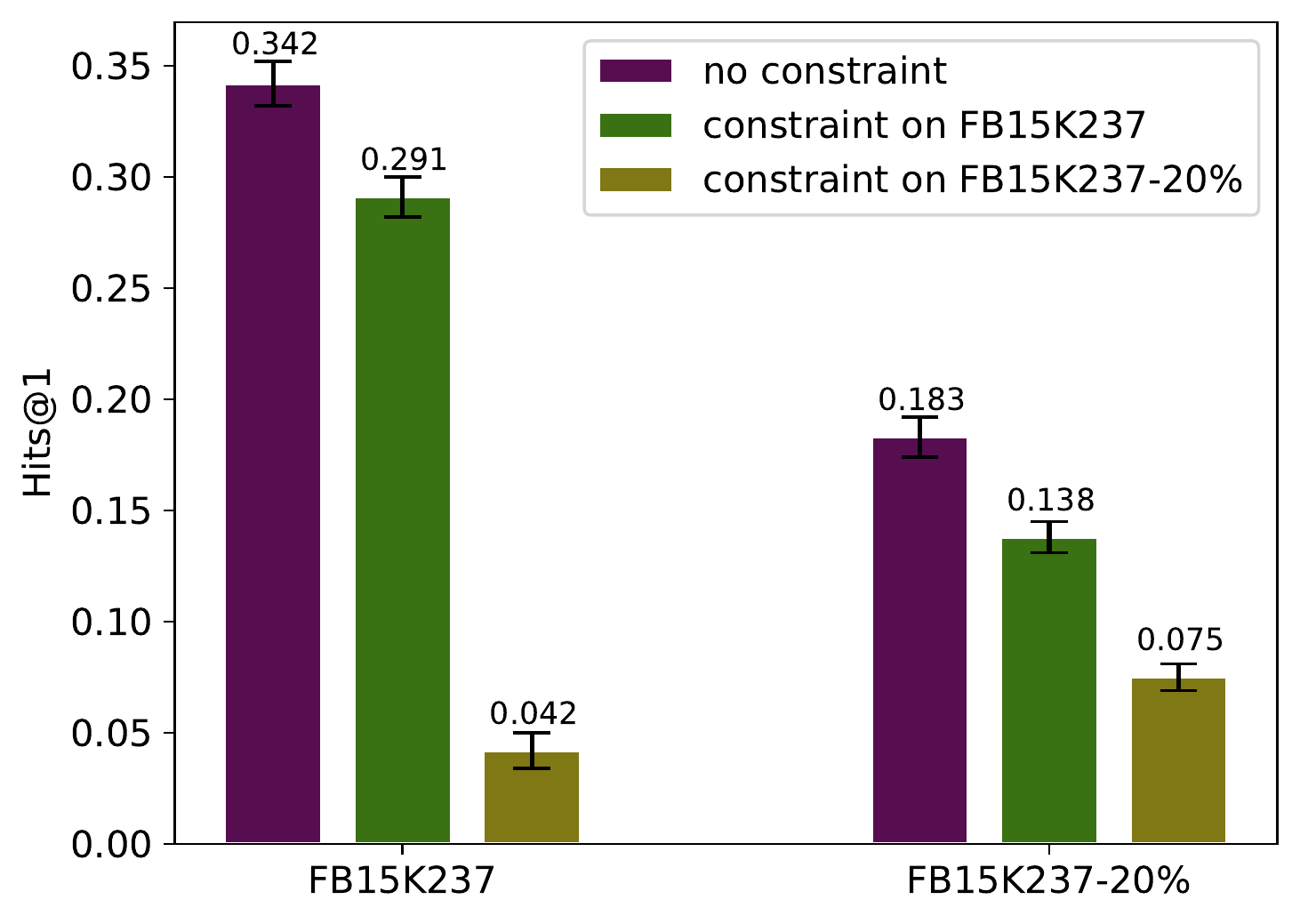}
\caption{Models trained on FB15K237 and FB15K237-20\% are evaluated on the same valid set. Different bars correspond to Hits@1 under different constraints during generation.}
\label{fig:sparse}
\end{figure}

We study the effect of constraint on our model's performance on link prediction.
Under the constraint of a set of edges, we treat a path as a valid path if it only contains edges in the constraint, and evaluate on all valid paths that the model generates.
In Fig.\ref{fig:sparse}, we report Hits@1 of models trained on FB15K237 and its subgraph FB15K237-20\%\footnote{The comparison is fair, since FB15K237 and FB15K237-20\% share the same valid/test set.}, under constraints of edges in FB15K237 and edges in FB15K237-20\%.

We see that for the model trained on FB15K237, out of 34.2\% paths that it correctly predicted (without constraint), 85\% of them contain only edges from the trained graph (0.342 $\rightarrow$ 0.291).
This indicates that in a more complete graph, most of the generated paths exist in the graph.
Meanwhile, for the model trained on FB15K237-20\%, only 41\% of all Hits@1 paths contain only edges from the trained graph (0.183 $\rightarrow$ 0.075), while 75\% of them contain only edges from FB15K237 (0.183 $\rightarrow$ 0.138), a more complete set of edges.
In other words, at least 34\% of all Hits@1 paths are inferred by dynamically completing missing edges along the path.
The results suggest that our model can indeed, ``walk and complete'', which has been shown to be effective in sparse setting.

\section{Conclusion}
\label{sec:conclusion}

This paper introduces SQUIRE, a sequence-to-sequence framework for efficient and effective multi-hop knowledge graph reasoning.
SQUIRE treats the triple query and the evidential path as sequences and utilizes Transformer to learn and infer in an end-to-end fashion.
We propose rule-enhanced learning and iterative training to further boost performance.
Experiments show that our approach significantly outperforms previous methods, on both standard KGs and sparse KGs.
Compared with RL-based methods, SQUIRE has faster convergence, and can be efficiently trained on larger KGs.
Moreover, we show that reasoning paths inferred by SQUIRE are more convincing than those generated by RL-based baselines.

\section*{Limitations}
Although our model generates more reasonable paths, as Table~\ref{tb:interpretability} suggests, the interpretability score and reasonable rate are still low for practical concerns.
We recognize this problem as a lack of high-quality KG datasets for multi-hop reasoning.
Since, as we observed, there is no reasonable path in the graph for more than 70\% of the triple queries in FB15K237, due to the missing of relevant nodes and edges in the graph.
In future work, it is worthwhile to construct KG datasets that provide more available reasonable paths to facilitate studies on interpretable multi-hop KG reasoning.

\section*{Acknowledgement}
This work is supported by the NSFC Youth Project (62006136), the Institute for Guo Qiang, Tsinghua University (2019GQB0003) and the grant from Alibaba Inc.

\bibliography{anthology}
\bibliographystyle{acl_natbib}

\appendix

\appendix
\onecolumn

\section{Training Details}
\label{sec:detail}

\begin{table}[htbp]
\centering  
\resizebox{0.65\textwidth}{!}{
\begin{tabular}{ccccccc}
\toprule[2pt]
Dataset & $lr$ & $\epsilon$ & $p$ & $\alpha$ & epoch & beam size \\
\midrule
FB15K237 & 0.0005 & 0.25 & 0.15 & 1/3 & 30 & 256 \\
NELL995 & 0.001 & 0.25 & 0.15 & 1/10 & 30 & 512 \\
FB15K237-20\%  & 0.0001 & 0.25 & 0.25 & 1/3 & 30 & 256 \\
NELL23K & 0.0005 & 0.25 & 0.15 & 1/3 & 100 & 512 \\
KACC-M & 0.001 & 0.75 & 0.15 & 1/10 & 20 & 256 \\
FB100K & 0.001 & 0.55 & 0.15 & 1/3 & 40 & 256 \\
\bottomrule[2pt]
\end{tabular}
}
\caption{Best hyperparameters on each dataset.}
\label{tb:hyper}
\end{table}
The Transformer Encoder used in our model contains 6 Transformer encoder layers. All layers have embedding size $d$ of 256, feedforward dimension of 512, 4 attention heads, and dropout of 0.1.
We report the best hyperparameters of SQUIRE on each dataset in Table~\ref{tb:hyper}.
These hyperparameters include learning rate $lr$, label smoothing factor $\epsilon$, token mask probability $p$, ratio of warmup step $\alpha$\footnote{Warmup for learning rate schedule is vital for training Transformer model~\cite{vaswani2017attention}, where the learning rate increase from 0 to its peak during warmup steps and slowly decrease to 0 afterward. $\alpha$ is the ratio of the number of warmup steps over the total optimization steps.}, number of epochs and beam size during beam searching.

On all datasets, we set maximum hops of path $N=3$, in other words, we only consider evidential path that contains 3 or fewer hops (since longer paths may not be as meaningful, especially in dense KGs).
To diversify the training set, we sample 6 query-path pairs from each triple.
This means that for rule-enhanced learning, we obtain 6 different paths from 6 logical rules, where rules with higher confidence scores come first.
On larger datasets (KACC-M and FB100K), we turn off iterative training strategy since its time cost is large.

\section{Convergence Analysis}
\label{sec:convergence}
\begin{figure}[htbp]
\centering
\includegraphics[width=0.5\linewidth,trim=0 0 0 0,clip]{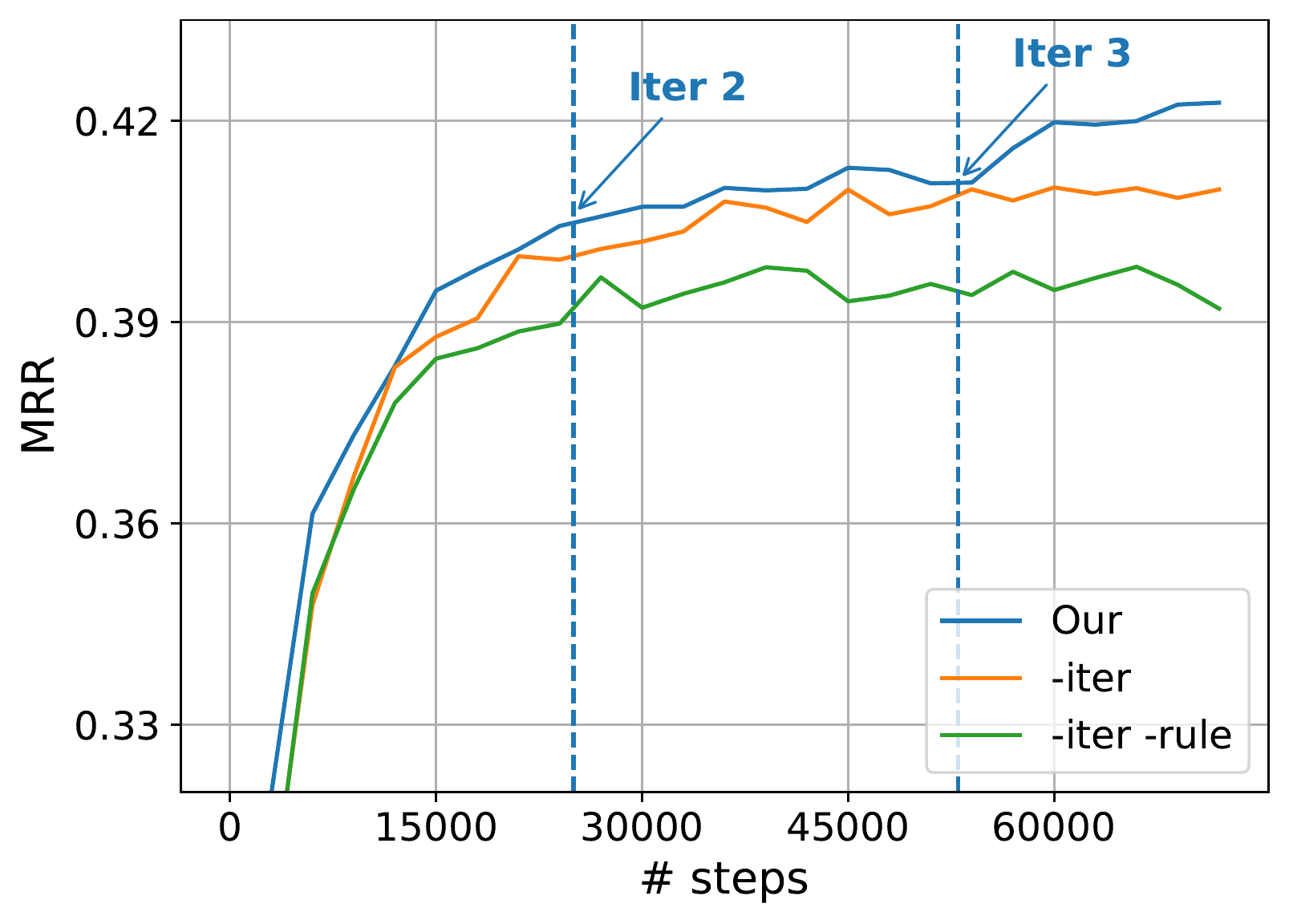}
\caption{Ablation results on convergence rate, on FB15K237 dataset. The curves show the valid set MRR w.r.t the number of optimization steps. The blue curve represents our original SQUIRE model, while the orange and green curves represent two ablated models.}
\label{fig:ablation}
\end{figure}

We compare the convergence rate of the original model with the two ablated models trained on FB15K237 dataset.
The convergence result is shown in Fig.~\ref{fig:ablation}, the curves show the trend in test MRR w.r.t. the number of optimization steps (all other hyperparameters are kept the same for the three models).
We observe that under iterative training, our model gains a boost on MRR at the start of each iteration, i.e., after data aggregation with the current model, as suggested by the trend of the blue curve at the start of Iter 2 and Iter 3 in the figure.
This indicates that the newly aggregated data during training helps the model's overall performance.
Furthermore, we can see that the convergence curve is more fluctuating without rule-enhanced learning (green curve).
This instability comes from the noise in random sampling, where many paths in the training set may have been meaningless and misleading.
On the other hand, rule-guided searching improves the quality of the paths in training set thus resulting in a more stable convergence.

\begin{table*}[t]
\centering
\resizebox{1\textwidth}{!}{
\begin{tabular}{lc}
\toprule[2pt]
Query-path pairs                                     & \multicolumn{1}{l}{score} \\
\midrule
Query: \textit{(Grammy Award for Best Long Form Music Video, award ceremony, ?)} & \multirow{2}{*}{1.0}        \\
Path: \textit{Grammy Award for Best Long Form Music Video} $\xrightarrow{\textit{category of}}$ \textit{Grammy Award} $\xrightarrow{\textit{instance of event}}$ \textit{51st Grammy Awards} & \\
\midrule
Query: \textit{(Concordia University, located in, ?)} & \multirow{2}{*}{0.5}        \\
Path: \textit{Concordia University} $\xleftarrow{\textit{contains}}$ \textit{Montreal} $\xrightarrow{\textit{contains}}$ \textit{McGill University} $\xleftarrow{\textit{contains}}$ \textit{Quebec} & \\
\midrule
Query: \textit{(Arthur Wellesley, gender, ?)} & \multirow{2}{*}{0}        \\
Path: \textit{Arthur Wellesley} $\xrightarrow{\textit{official position}}$ \textit{Prime minister} $\xleftarrow{\textit{official position}}$ \textit{Pierre Trudeau} $\xrightarrow{\textit{gender}}$ \textit{Male} & \\
\bottomrule[2pt]
\end{tabular}
}
\caption{Several annotation examples for query-path pairs with scores of 1.0, 0.5 and 0. The score is given based on to what degree the path can convince a human that the final entity along the path is the answer to the query.}
\label{tb:annotation}
\end{table*}

\section{Interpretability Annotation }
\label{sec:case}
To evaluate whether the model's generated multi-hop paths are convincing to a human, we randomly choose 100 triple queries from test set and obtain the top path generated for each query $(h, r)$ that reaches the target entity $t$.
Then we score these paths into three grades: reasonable paths are given 1.0, partially reasonable paths are given 0.5 and unreasonable paths are given 0\footnote{Note that if none of the generated paths reach the correct target entity, we also give 0 score for that query.}. 
Here is how we determine the score of each path:
\begin{enumerate}
    \item Reasonable paths are those that can sufficiently deduce the relationship $r$ between $h$ and $t$.
    \item Partially reasonable paths are those that cannot sufficiently deduce the relationship, but they are in highly positive correlation with the triple fact.
    \item Unreasonable paths are those that are irrelevant to the triple fact.
\end{enumerate}
We give several annotation examples in Table~\ref{tb:annotation}.
The full annotation table can be found \href{https://docs.google.com/spreadsheets/d/1pLtOe1HUYAgs41HWZwBpA3ahHlW0GM01KoH92seFzGg/edit?usp=sharing}{here}.

\begin{figure*}[t]
\centering
\includegraphics[width=0.9\linewidth,trim=0 40 0 0,clip]{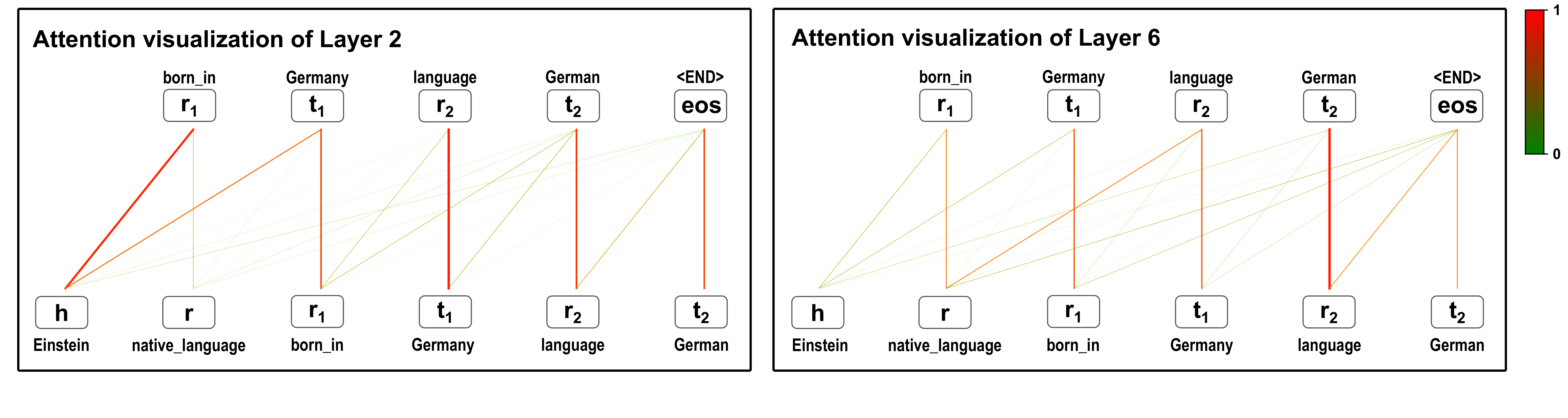}
\caption{Attention visualization of Transformer encoder layer 2 and 6 in our trained model. 
Colored lines show the attention from tokens in the predicted path (top line) to the input query and their previous tokens (bottom line), recall that on each position, the Transformer encoder can only attend to query tokens and its previous tokens.
Thicker line indicates a higher degree of attention. More straightforwardly, the attention value varies from 0 to 1, with color changing from green to red. }
\label{fig:attn}
\end{figure*}

\section{Attention Analysis}
\label{sec:att}
Having seen the promising performance of SQUIRE, we want to find out more about how our sequence-to-sequence model infers the evidential path. 
To this end, we visualize the attention matrix in each Transformer encoder layer for the query \texttt{(Einstein, native\_language, ?)} along with its evidential path.
The attention visualization of layer 2 and layer 6 are shown in Fig.~\ref{fig:attn}.
For each token in the predicted path, we observe massive attention paid to its predecessor token, i.e., the previous step along the path. 
For example, $r_1$ on $h$, $t_1$ on $r_1$, $r_2$ on $t_1$ and so on.
This explains how SQUIRE generates a valid path without any supervision about the graph: the model has memorized the edges during training, and in generation phase it predicts the next step based on the current position and adjacent edges or nodes.

Moreover, comparing the two visualizations, we find that the latter layer aggregates more global information during self-attention computation.
Particularly, in layer 6, there is a moderate level of attention from $r_1, r_2$ to query relation $r$.
This suggests that SQUIRE may distinguish the real evidential path from other spurious paths by attending to the query.

\end{document}